\documentclass[journal]{IEEEtran}
\usepackage{graphicx}
\usepackage{subfig}
\usepackage[compress]{cite}
\usepackage{url}
\usepackage{paralist}
\usepackage{algorithm}
\usepackage{algorithmic}
\usepackage{hyperref}
\hypersetup{
    bookmarks=true,         
    unicode=false,          
    pdftoolbar=true,        
    pdfmenubar=true,        
    pdffitwindow=false,     
    pdfstartview={FitH},    
    pdftitle={My title},    
    pdfauthor={Author},     
    pdfsubject={Subject},   
    pdfcreator={Creator},   
    pdfproducer={Producer}, 
    pdfkeywords={keyword1, key2, key3}, 
    pdfnewwindow=true,      
    colorlinks=true,       
    linkcolor=black,          
    citecolor=black,        
    filecolor=black,      
    urlcolor=black           
}
\ifCLASSINFOpdf
\else
\fi
\begin{document}
%
\title{Supervised Discrete Hashing with Relaxation}
%
%
%

\author{Jie~Gui$^*$,~\IEEEmembership{Senior Member,~IEEE,}
        Tongliang~Liu$^*$,
        Zhenan~Sun,~\IEEEmembership{Member,~IEEE,}
        Dacheng~Tao,~\IEEEmembership{Fellow,~IEEE,}
        and~Tieniu~Tan,~\IEEEmembership{Fellow,~IEEE}
\thanks{This work was supported in part by the grant of the National Science Foundation of China under Grant 61572463, Grant 31371340, in part by CCF-Tencent Open Fund, and in part by Australian Research Council Projects DP-140102164, FT-130101457, and LE-140100061. All correspondence should be directed to Zhenan Sun.
\it{(Jie Gui and Tongliang Liu contributed equally to this work.)}

\rm{J. Gui is with Institute of Intelligent Machines, Chinese Academy of Sciences, Hefei 230031, People's Republic of China. E-mail: guijie@ustc.edu.}}
\thanks{T. Liu is with the Centre for Artificial Intelligence and the Faculty of Engineering and Information Technology, University of Technology Sydney, Sydney, NSW 2007, Australia. E-mail: tliang.liu@gmail.com.}
\thanks{D. Tao is with School of Information Technologies, Faculty of Engineering and Information Technologies, University of Sydney, NSW, Australia . E-mail: dacheng.tao@gmail.com}
\thanks{ Z. Sun and T. Tan are with the Center for Research on Intelligent Perception and Computing, National Laboratory of Pattern Recognition, Institute of Automation, CAS Center for Excellence in Brain Science and Intelligence Technology, Chinese Academy of Sciences, Beijing, 100190, China. E-mail: \{znsun, tnt\}@nlpr.ia.ac.cn.}
\thanks{$^*$ }
\thanks{ \textcopyright 20XX IEEE. Personal use of this material is permitted. Permission from IEEE must be obtained for all other uses, in any current or future media, including reprinting/republishing this material for advertising or promotional purposes, creating new collective works, for resale or redistribution to servers or lists, or reuse of any copyrighted component of this work in other works.¡±}
}

%
%

\markboth{IEEE Transactions on Neural Networks and Learning Systems}%
{Shell \MakeLowercase{\textit{et al.}}: Bare Demo of IEEEtran.cls for Journals}
%



\maketitle

\begin{abstract}
Data-dependent hashing has recently attracted attention due to being able to support efficient retrieval and storage of high-dimensional data such as documents, images, and videos. In this paper, we propose a novel learning-based hashing method called ``Supervised Discrete Hashing with Relaxation" (SDHR) based on ``Supervised Discrete Hashing" (SDH). SDH uses ordinary least squares regression and traditional zero-one matrix encoding of class label information as the regression target (code words), thus fixing the regression target. In SDHR, the regression target is instead optimized. The optimized regression target matrix satisfies a large margin constraint for correct classification of each example. Compared with SDH, which uses the traditional zero-one matrix, SDHR utilizes the learned regression target matrix and, therefore, more accurately measures the classification error of the regression model and is more flexible. As expected, SDHR generally outperforms SDH. Experimental results on two large-scale image datasets (CIFAR-10 and MNIST) and a large-scale and challenging face dataset (FRGC) demonstrate the effectiveness and efficiency of SDHR.
\end{abstract}


%
\IEEEpeerreviewmaketitle

\section{Introduction}
%
%
%
%
\IEEEPARstart{I}{n} large-scale visual searching, data must be indexed and organized accurately and efficiently. Hashing has attracted the interest of researchers in machine learning, information retrieval, computer vision, and related communities and has shown promise for large-scale visual searching. This paper focuses on hashing algorithms that encode images, videos, documents, or other data types as a set of short binary codes while preserving the original example structure (e.g., similarities between data points). One advantage of these hashing methods is that pairwise distance calculations can be performed extremely efficiently in Hamming space and, as a result, methods based on pairwise comparisons can be performed more efficiently and applied to large datasets. Due to the flexibility of binary representations, hashing algorithms can be applied in many ways, for example, searching efficiently by exploring only examples falling into buckets close to the query according the Hamming distance or using the hash codes for other tasks such as image classification, face recognition, and indexing.

Hashing methods can be broadly classified into two categories: data-independent and data-dependent methods. Data-independent algorithms do not require training data and randomly construct a set of hash functions without any training. Representative methods include locality-sensitive hashing (LSH) \cite{gionis1999similarity} and its variants \cite{raginsky2009locality} and the Min-Hash algorithms \cite{chum2010large}. A disadvantage of the LSH family is that they usually need a long bit length ($\ge$1000) for satisfactory performance. This results in large storage costs, thus limiting their applications.

Recently, data-dependent or learning-based hashing methods have become popular because learned compact hash codes can effectively and efficiently index and organize massive amounts of data. Instead of randomly generating hash functions as in LSH, data-dependent hashing methods aim to generate short hash codes (typically $\le $200) using training data. There is abundant literature on different data-dependent hashing methods that can be classified into four main categories.

The first is unsupervised hashing, which does not utilize the label information of training examples. Representative algorithms include spectral hashing (SH) \cite{weiss2009spectral}, principal component hashing (PCH) \cite{matsushita2009principal}, principal component analysis \cite{hu2016sparse} iterative quantization (PCA-ITQ) \cite{gong2013iterative,gong2011iterative}, anchor graph-based hashing (AGH) \cite{liu2011hashing}, scalable graph hashing with feature transformation (SGH) \cite{jiang2015scalable}, and inductive manifold hashing (IMH) \cite{shen2013inductive} with t-distributed stochastic neighbor embedding (t-SNE) \cite{maaten2008visualizing}. However, since unsupervised hashing does not consider the label information of input data, useful information critical for pattern classification may be lost. Thus, various supervised hashing and semi-supervised hashing techniques have been proposed since it is generally believed that label information produces a more discriminative recognition algorithm.

The second category is supervised and semi-supervised hashing, which take full consideration of the class labels. Representative methods in this group include semi-supervised hashing (SSH) \cite{wang2012semi}, kernel-based supervised hashing (KSH) \cite{liu2012supervised}, fast supervised hashing using graph cuts and decision trees (FastHash) \cite{lin2015supervised,lin2014fast}, supervised discrete hashing (SDH) \cite{shen2015supervised}, and linear discriminant analysis (LDA) \cite{zhangsparse} based hashing (LDAHash) \cite{strecha2012ldahash}. In our view, ranking-based methods \cite{norouzi2012hamming,wang2013order,wang2013learning,wang2015ranking} (in which ranking labels such as triplets constitute the supervised information) also belong to the supervised hashing group.

The third category is multimodal hashing, which includes multi-source hashing and cross-modal hashing. Multi-source hashing \cite{song2011multiple,zhang2011composite} assumes that all the views are provided for a query and aims to learn better codes than unimodal hashing by using all these views. In cross-modal hashing \cite{kumar2011learning}, the query represents one modality while the output represents another modality. For example, given a text query, images are returned corresponding to the text. Therefore, both multi-source hashing and cross-modal hashing use multi-modal information \cite{chao2015alternative}. However, they are used in different applications and cross-modal hashing may have wider application than multi-source hashing in practice.

The fourth category is deep learning \cite{Gong2016Change,Xia2016Bottom,Dundar2016Embedded} based hashing. To the best of our knowledge, semantic hashing \cite{Salakhutdinov2009} represents the first use of deep learning for hashing. This seminal work utilized the stacked-restricted Boltzmann machine (RBM) to learn compact hash codes for visual searching. However, the model was complex and required pre-training, which is inefficient in practice. Both deep regularized similarity comparison hashing (DRSCH) \cite{zhang2015bit} and \cite{lin2015deep} use the deep convolutional neural network (CNN) for hashing. Some other related methods can be found in \cite{lai2015simultaneous,li2015feature,liudeep}.

In general, the discrete constraints imposed on the hash codes generated by the hash objective functions lead to NP-hard mixed-integer optimization problems. To simplify the optimization in the hash code learning process, most algorithms first solve a relaxed problem by discarding the discrete constraints and then perform a quantization step that turns real values into the approximate hash code by thresholding (or quantization). This relaxation strategy greatly simplifies the original discrete optimization. However, such an approximate solution is suboptimal, typically of low quality, and often produces a less effective hash code due to the accumulated quantization error, especially when learning long-length codes. Most existing hashing methods do not take discrete optimization into account. Shen et al. \cite{shen2015supervised} proposed a novel Supervised Discrete Hashing (SDH) method that aimed to directly learn the binary hash codes without relaxations. To make full use of label information, this method was formulated as a least squares classification to regress each hash code to its corresponding label.

However, the ordinary least squares problem may not be optimal for classification. To further improve the performance, here we propose a novel method called ``Supervised Discrete Hashing with Relaxation" (SDHR) to directly learn the regression targets from data. SDHR is essentially a single and compact learning method for multiclass classification. The regression targets learned by SDHR can guarantee that each example (data point) is correctly classified with a large margin. During learning, SDHR does not consider the absolute values in regression targets and only forces relative values to satisfy a large margin for correct classification. Therefore, SDHR is much more accurate and flexible than SDH. The optimization problem of solving the regression target for SDHR is convex, and we employ an efficient alternating procedure to find the regression target. Our experimental results show that SDHR generally performs better than SDH.

The remainder of this paper is organized as follows. Section II outlines SDH. In Section III, we describe our proposed method in detail. The experimental results are presented in Section IV, followed by conclusions in Section V.

\section{Brief review of supervised discrete hashing}
In this section, we introduce the related works SDH \cite{shen2015supervised} by way of introduction. Suppose that we have $n$ examples $X = \{ {x_i}\} _{i = 1}^n$ (Table \ref{tab:0} for a list of notations used in this paper). We aim to learn a set of hash codes $B = \{ {b_i}\} _{i = 1}^n \in {\{  - 1,1\} ^{n \times l}}$ to preserve their similarities in the original space, where the $i$th vector ${b_i}$ is the $l$-bits hash codes for ${x_i}$. The corresponding class labels of all training examples are denoted as $Y = \{ {y_i}\} _{i = 1}^n \in {R^{n \times c}}$, where $c$ is the number of classes and ${y_{ik}} = 1$ if ${x_i}$ belongs to class $k$ and 0 otherwise. The term ${y_{ik}}$ is the $k$th element of ${y_i}$.
\begin{table}
\caption{Notations} \label{tab:0}
\begin{center}
\begin{tabular}{|c|c|}\hline
 Notation & Description\\
\hline $X$ & the data matrix
\\ ${x_i}$ & the $i$-th data point
\\ $n$ & the training sample size: the number of the total training data points
\\ $B$ & the hash codes
\\ ${b_i}$ & the $i$-th row of $B$ (the hash code for ${x_i}$)
\\ $l$ &the length of hash code
\\ $Y$ &the label matrix
\\ $c$ &the number of classes
\\ ${y_i}$ &the $i$-th row of the matrix $Y$
\\ ${y_{ik}}$ & the $k$-th element of ${y_i}$
\\ $W$ & the projection matrix for the hash code
\\ $F\left(  \cdot  \right)$ & a nonlinear embedding to approximate the hash code
\\ $m$ & the number of anchor points
\\ $\phi \left(  \cdot  \right)$ & an $m$-dimensional row vector obtained by the RBF kernel
\\ $P$ & the projection matrix for the nonlinear embedding
\\ $t$ & the translation (offset) vector used in SDHR
\\ $R$ & the regression target matrix (code words) used in SDHR
\\ ${R_i}$ & the $i$-th row of the matrix $R$
\\ ${L_i}$ & the label of ${x_i}$
\\ ${e_n}$ & an $n$-dimensional column vector with all elements equal to one
\\ $r$ & the Hamming radius
\\ \hline
\end{tabular}
\end{center}
\end{table}

The objective function of SDH is defined as:
\begin{eqnarray}\label{equ:1}
\begin{array}{l}
 \mathop {\min }\limits_{B,F,W} \sum\limits_{i = 1}^n {\left\| {{y_i} - {b_i}W} \right\|_2^2}  + \lambda \left\| W \right\|_F^2 \\
 \quad \quad \quad \quad  + v\sum\limits_{i = 1}^n {\left\| {{b_i} - F\left( {{x_i}} \right)} \right\|_2^2}  \\
 s.t.\;\forall i\quad {b_i} \in {\left\{ { - 1,1} \right\}^l}. \\
 \end{array}
\end{eqnarray}
That is,
\begin{eqnarray}\label{equ:2}
\begin{array}{l}
 \mathop {\min }\limits_{B,F,W} \left\| {Y - BW} \right\|_F^2 + \lambda \left\| W \right\|_F^2 + v\left\| {B - F\left( X \right)} \right\|_F^2 \\
 s.t.\;B \in {\left\{ { - 1,1} \right\}^{n \times l}}, \\
 \end{array}
 \end{eqnarray}
 where  ${\left\|  \cdot  \right\|_F}$  is the Frobenius norm of a matrix. The first term of (\ref{equ:1}) is the ordinary least squares regression, which is used to regress each hash code to its corresponding class label. The term $W$ is the projection matrix. The second term of (\ref{equ:1}) is for regularization. $F\left(  \cdot  \right)$ in the last term of (\ref{equ:1}) is a simple yet powerful nonlinear embedding to approximate the hash code
\begin{eqnarray}\label{equ:3}
F\left( x \right) = \phi \left( x \right)P,
\end{eqnarray}
where $\phi \left( x \right)$ is an $m$-dimensional row vector obtained by the RBF kernel: $\phi \left( x \right) = [\exp \left( {{{{{\left\| {x - {a_1}} \right\|}^2}} \mathord{\left/
 {\vphantom {{{{\left\| {x - {a_1}} \right\|}^2}} \sigma }} \right.
 \kern-\nulldelimiterspace} \sigma }} \right), \cdots ,\exp \left( {{{{{\left\| {x - {a_m}} \right\|}^2}} \mathord{\left/
 {\vphantom {{{{\left\| {x - {a_m}} \right\|}^2}} \sigma }} \right.
 \kern-\nulldelimiterspace} \sigma }} \right)]$, $\{ {a_j}\} _{i = 1}^m$ are randomly selected $m$ anchor points from the training examples, and $\sigma$ is the Gaussian kernel parameter. The matrix $P \in {R^{m \times l}}$ projects $\phi \left( x \right)$ onto the low-dimensional space. Similar formulations as equation (\ref{equ:3}) are widely used in other methods such as kernel-based supervised hashing (KSH) \cite{liu2012supervised} and binary reconstructive embedding (BRE) \cite{kulis2009learning}.

 The need of a Gaussian kernel function is shown as follows: existing methods such as locality-sensitive hashing (LSH) do not apply for high-dimensional kernelized data when the underlying feature embedding for the kernel is unknown. Moreover, the using of kernel generalizes such methods as locality-sensitive hashing to accommodate arbitrary kernel functions, making it possible to preserve the algorithm's sublinear time similarity search guarantees for a wide class of useful similarity functions.

The optimization of (\ref{equ:2}) involves three steps: the F-step solving $P$, the G-step solving $W$, and the B-step solving $B$:

\textbf{F-step}  By fixing $B$, the projection matrix $P$ is easily computed:
\begin{eqnarray}
P = {\left( {\phi {{\left( X \right)}^T}\phi \left( X \right)} \right)^{ - 1}}\phi {\left( X \right)^T}B.
\end{eqnarray}

\textbf{G-step}  If $B$ is fixed, it is easy to solve $W$, which has a closed-form solution:
\begin{eqnarray}
W = {\left( {{B^T}B + \lambda I} \right)^{ - 1}}{B^T}Y.
\end{eqnarray}

\textbf{B-step}  By fixing other variables, $B$ also has a closed-form solution. Please refer to \cite{shen2015supervised} for details.
\section{Our proposed method: Supervised Discrete Hashing with Relaxation}
In this section, we introduce our proposed SDHR method in detail. The first term of SDH's objective function is to regress ${y_i}$ on ${b_i}$. That is $\min \left\| {{y_i} - {b_i}W} \right\|_2^2$, and the optimal solution ${W^*}$ is ${y_i} = {b_i}{W^*}$, which is a linear function that will go through the origin of the coordinate. It is more appropriate and flexible to have a translation (offset) vector $t$ and employing the hypothesis ${y_i} = {b_i}W + {t^T}$, where ${t^T}$ is the transpose of the column vector $t$. In SDH, ${y_i}$ is fixed and ${y_{ij}} = 1$ if ${x_i}$ belongs to the class $j$ and 0 otherwise. To make SDH more discriminative and flexible, we use ${R_i}$ instead of ${y_i}$ as the regression target (code words) for the $i$th hash code ${b_i}$, which satisfies
\begin{eqnarray}
\begin{array}{l}
 \mathop {\min }\limits_{{R_i}} \left\| {{R_i} - {b_i}W - {t^T}} \right\|_F^2 \\
 s.t.\;{R_{i,{L_i}}} - \mathop {\max }\limits_{k \ne {L_i}} {R_{ik}} \ge 1, \\
 \end{array}
\end{eqnarray}
where ${L_i}$ is the label of ${x_i}$. The term ${R_i}$ is optimized instead of being given. Our aim is to produce a large margin between the code words of the true label and all the other wrong labels that is larger than one to satisfy the large margin criterion ${R_{i,{L_i}}} - \mathop {\max }\limits_{k \ne {L_i}} {R_{ik}} \ge 1$. The objective function of SDHR is therefore defined as
\begin{eqnarray}
\begin{array}{l}
 \mathop {\min }\limits_{B,R,t,F,W} \sum\limits_{i = 1}^n {\left\| {{R_i} - {b_i}W - {t^T}} \right\|_2^2}  + \lambda \left\| W \right\|_F^2 \\
 \quad \quad \quad \quad \quad  + v\sum\limits_{i = 1}^n {\left\| {{b_i} - F\left( {{x_i}} \right)} \right\|_2^2}  \\
 s.t.\;\forall i\quad {b_i} \in {\left\{ { - 1,1} \right\}^l} \\
 \quad \forall i\quad {R_{i,{L_i}}} - \mathop {\max }\limits_{k \ne {L_i}} {R_{ik}} \ge 1. \\
 \end{array}
\end{eqnarray}
That is,
\begin{eqnarray}\label{equ:8}
\begin{array}{l}
 \mathop {\min }\limits_{B,R,t,F,W} \left\| {R - BW - {e_n}{t^T}} \right\|_F^2 + \lambda \left\| W \right\|_F^2 \\
 \quad \quad \quad \quad \quad  + v\left\| {B - F\left( X \right)} \right\|_F^2 \\
 s.t.\;B \in {\left\{ { - 1,1} \right\}^{n \times l}} \\
 \quad \forall i\quad {R_{i,{L_i}}} - \mathop {\max }\limits_{k \ne {L_i}} {R_{ik}} \ge 1, \\
 \end{array}
\end{eqnarray}
where ${e_n}$ is an $n$-dimensional column vector with all elements equal to one. Hashing is not specifically designed for classification. Nevertheless, the supervised information is still very helpful. We want the hash codes to be similar for examples from the same class and we want them to be different for examples from different classes. Therefore, we relax the fixed label values to relative ones to satisfy the large margin criterion, which is more flexible than using the fixed label values.

The problem in (\ref{equ:8}) is a mixed binary optimization problem. By fixing the other variables, solving a single variable is relatively easy. Based on this decomposition, an alternating optimization technique can be adopted to iteratively and efficiently solve this optimization problem. Each iteration alternatively solves $B,\;R,\;W,\;t,\;P$. Therefore, SDHR's optimization has five steps:

\textbf{$t$-step} When all variables except $t$ are fixed, (\ref{equ:8}) can be equivalently rewritten as:
\begin{eqnarray}\label{equ:9}
\begin{array}{l}
 \mathop {\min }\limits_t \;\left\| {{e_n}{t^T} - \left( {R - BW} \right)} \right\|_2^2 \\
  = \mathop {\min }\limits_t \;tr\left( \begin{array}{l}
 \left( {te_n^T - {{\left( {R - BW} \right)}^T}} \right) \\
  \times \left( {{e_n}{t^T} - \left( {R - BW} \right)} \right) \\
 \end{array} \right). \\
 \end{array}
\end{eqnarray}
By setting the derivative of (\ref{equ:9}) with respect to $t$ to zero, $t$ can be solved with closed-form solution
\begin{eqnarray}\label{equ:10}
t = \frac{{{{\left( {R - BW} \right)}^T}{e_n}}}{n}.
\end{eqnarray}

\textbf{B-step} For solving $B$, (\ref{equ:8}) can be rewritten as:
\[\begin{array}{l}
 \mathop {\min }\limits_B \left\| {R - BW - {e_n}{t^T}} \right\|_F^2 + v\left\| {B - F\left( X \right)} \right\|_F^2 \\
  = \mathop {\min }\limits_B \left\| {BW - \left( {R - {e_n}{t^T}} \right)} \right\|_F^2 + v\left\| {B - F\left( X \right)} \right\|_F^2 \\
  = \mathop {\min }\limits_B \left\| {BW} \right\|_F^2 \\
 \quad \quad \quad \quad  - 2tr\left( {B\left( {W{{\left( {R - {e_n}{t^T}} \right)}^T} + vF{{\left( X \right)}^T}} \right)} \right) \\
 \quad \quad \quad \quad  + vtr\left( {{B^T}B} \right). \\
 \end{array}\]
Since $tr\left( {{B^T}B} \right)$ is a constant, we have
\begin{eqnarray}\label{equ:11}
\mathop {\min }\limits_B \left\| {BW} \right\|_F^2 - 2tr\left( {BQ} \right)\quad s.t.\;B \in {\left\{ { - 1,1} \right\}^{n \times l}},
\end{eqnarray}
where $Q = {\left( {(R - {e_n}{t^T}){W^T} + vF\left( X \right)} \right)^T}$.

Thus, $B$ can be solved using the discrete cyclic coordinate descent method, which is similar to solving $B$ in SDH.

\textbf{F-step}  The F-step of SDHR is the same as that of SDH:
\begin{eqnarray}\label{equ:12}
P = {\left( {\phi {{\left( X \right)}^T}\phi \left( X \right)} \right)^{ - 1}}\phi {\left( X \right)^T}B.
\end{eqnarray}

\textbf{G-step}  If all variables except $W$ are fixed, we put (\ref{equ:10}) into (\ref{equ:8}) and get
\begin{eqnarray}\label{equ:13}
\begin{array}{l}
 \mathop {\min }\limits_W \left\| {R - BW - \frac{{{e_n}e_n^T\left( {R - BW} \right)}}{n}} \right\|_F^2 + \lambda \left\| W \right\|_F^2 \\
  = \mathop {\min }\limits_W \left\| {\left( {I - \frac{{{e_n}e_n^T}}{n}} \right)\left( {R - BW} \right)} \right\|_F^2 + \lambda \left\| W \right\|_F^2. \\
 \end{array}
\end{eqnarray}
By setting the derivative of (\ref{equ:13}) with respect to $W$ to zero, it is easy to solve $W$, which has a closed-form solution:
\begin{eqnarray}\label{equ:14}
W = {\left( {{B^T}\left( {I - \frac{{{e_n}e_n^T}}{n}} \right)B + \lambda I} \right)^{ - 1}}{B^T}\left( {I - \frac{{{e_n}e_n^T}}{n}} \right)R.
\end{eqnarray}

\textbf{R-step}  When we fix $B,\;t,\;P$ and $W$, how to solve $R$ in (\ref{equ:8}) can be transformed to
\begin{eqnarray}\label{equ:15}
\begin{array}{l}
 \mathop {\min }\limits_R \left\| {R - \left( {BW + {e_n}{t^T}} \right)} \right\|_F^2 \\
 s.t.\;\forall i\quad {R_{i,{L_i}}} - \mathop {\max }\limits_{k \ne {L_i}} {R_{ik}} \ge 1. \\
 \end{array}
\end{eqnarray}

It is natural to solve the matrix $R$ row by row, which has a general form of:
\begin{eqnarray}\label{equ:16}
\begin{array}{l}
 \mathop {\min }\limits_r \sum\nolimits_{q = 1}^c {{{\left( {{r_q} - {a_q}} \right)}^2}}  \\
 s.t.\;{r_j} - \mathop {\max }\limits_{k \ne j} {r_k} \ge 1, \\
 \end{array}
\end{eqnarray}
where $r$ and $a$ are the $i$th row of $R$ and $\left( {BW + {e_n}{t^T}} \right)$, respectively; and ${r_q}$ and ${a_q}$ are the $q$th element of $r$ and $a$, respectively. We can use the Lagrangian multiplier method to solve (\ref{equ:16}). The Lagrangian function is defined as:
\begin{eqnarray}\label{equ:17}
\begin{array}{l}
 L\left( {{r_q},{\lambda _k}} \right) = \sum\nolimits_{q = 1}^c {{{\left( {{r_q} - {a_q}} \right)}^2}}  \\
 \quad \quad \quad \quad \quad \quad  + \sum\nolimits_{k = 1,k \ne j}^c {{\lambda _k}\left( {1 + {r_k} - {r_j}} \right)} , \\
 \end{array}
\end{eqnarray}
where ${\lambda _k}$ is the Lagrangian multiplier and ${\lambda _k} \ge 0$. By setting the derivative of $L\left( {{r_q},{\lambda _k}} \right)$ with respect to ${r_j}$ and ${r_k}\left( {k \ne j} \right)$ to zero, we have
\begin{eqnarray}\label{equ:18}
2\left( {{r_j} - {a_j}} \right) - \sum\nolimits_{k = 1,k \ne j}^c {{\lambda _k} = 0} ,
\end{eqnarray}
\begin{eqnarray}\label{equ:19}
2\left( {{r_k} - {a_k}} \right) + {\lambda _k} = 0,\;\quad k \ne j.
\end{eqnarray}
Furthermore, using the Kuhn-Tucker condition, we have:
\begin{eqnarray}\label{equ:20}
{\lambda _k}\left( {1 + {r_k} - {r_j}} \right) = 0,\quad \quad k \ne j.
\end{eqnarray}
With (\ref{equ:18}), (\ref{equ:19}), and (\ref{equ:20}), we have $\left( {2c - 1} \right)$ equations and $\left( {2c - 1} \right)$ variables. Thus, (\ref{equ:16}) can be solved in this way. Please note that (\ref{equ:16}) can be solved using other optimization methods. For example, a method called retargeting proposed in \cite{Zhang2015Retargeted} can also used to solve this optimization problem. In summary, our algorithm for solving SDHR is presented in Algorithm 1.
\begin{algorithm}
\caption{Supervised Discrete Hashing with Relaxation (SDHR)} \label{alg:FSSL1}
\begin{algorithmic}
\STATE {\textbf{Inputs:} training data $\{ {x_i},{y_i}\} _{i = 1}^n$; code length $l$; maximum iteration number $t$; parameter $\lambda $}
\STATE {\textbf{Output:} binary codes $\{ {b_i}\} _{i = 1}^n \in {\left\{ { - 1,1} \right\}^{n \times l}}$}
\STATE Randomly select $m$ examples $\{ {a_j}\} _{i = 1}^m$ from the training examples and get the $\phi \left( x \right)$ via the Gaussian kernel function;
\STATE Initialize ${b_i}$ as a ${\{  - 1,1\} ^l}$ vector randomly;
\STATE Initialize $R$ as $R = \left\{ {{R_{ij}}} \right\} \in {R^{n \times c}}$ where ${R_{ij}} = \left\{ \begin{array}{l}
 1,\;if\;{y_i} = j \\
 0,\;otherwise \\
 \end{array} \right.$;
\STATE Use (\ref{equ:14}) to initialize $W$;
\STATE Use (\ref{equ:10}) to initialize $t$;
\STATE Use (\ref{equ:12}) to initialize P;
\REPEAT
\STATE \textbf{B-step} Use (\ref{equ:11}) to solve $B$;
\STATE \textbf{R-step} Use (\ref{equ:18}), (\ref{equ:19}) and (\ref{equ:20}) to solve $R$;
\STATE \textbf{G-step} Use (\ref{equ:14}) to solve $W$;
\STATE \textbf{t-step} Use (\ref{equ:10}) to solve $t$;
\STATE \textbf{F-step} Use (\ref{equ:12}) to solve $P$;
\UNTIL{convergence}
\end{algorithmic}
\end{algorithm}

%



%
%
\section{Experimental results}
In this section, we investigate SDHR's performance by conducting experiments on a server with an Intel Xeon processor (2.80 GHz), 128GB RAM, and configured with Microsoft Windows Server 2008 and MATLAB 2014b.

We conduct experiments on two large-scale image datasets, CIFAR-10\footnote{\url{https://www.cs.toronto.edu/~kriz/cifar.html}} and MNIST\footnote{\url{http://yann.lecun.com/exdb/mnist/}}, and a large-scale and challenging face dataset FRGC. The proposed method is compared with popular hashing methods including BRE \cite{kulis2009learning}, SSH \cite{wang2012semi}, KSH \cite{liu2012supervised}, FastHash \cite{lin2015supervised,lin2014fast}, AGH \cite{liu2011hashing}, and IMH \cite{shen2013inductive} with t-SNE \cite{maaten2008visualizing}. For iterative quantization (ITQ) \cite{gong2013iterative,gong2011iterative}, we use both its supervised version CCA-ITQ and unsupervised version PCA-ITQ. Canonical correlation analysis (CCA) is used as the preprocessing step for CCA-ITQ. We use the public MATLAB codes and the parameters suggested by the corresponding authors. Specifically, for SDH and SDHR, $\lambda$ and $v$ are empirically set to 1 and 1e-5, respectively; we set the maximum iteration number $t$ to 5. For AGH, IMH, SDH, and SDHR, 1,000 randomly selected anchor points are used.

The experimental results are reported in terms of Hamming ranking (mean average precision, MAP), hash lookup (precision, recall, F-measure of Hamming radius 2), accuracy, training time, and test time. The Hamming radius $r$ is set to be 2 as in \cite{erin2015deep,shen2015supervised}. The F-measure is defined as 2$\times$precision$\times$recall/(precision + recall). The following evaluation metric is also utilized to measure the performance of the different algorithms: precision at $N$ samples (precision@sample = $N$), which is the percentage of true neighbors among the top $N$ retrieved instances. $N$ is set to be 500 as in \cite{erin2015deep}. Note that a query is considered to be a false case if no example is returned when calculating precisions. Ground truths are defined by the label information of the datasets.

\begin{figure}
\begin{center}
\scalebox{0.6}{\includegraphics{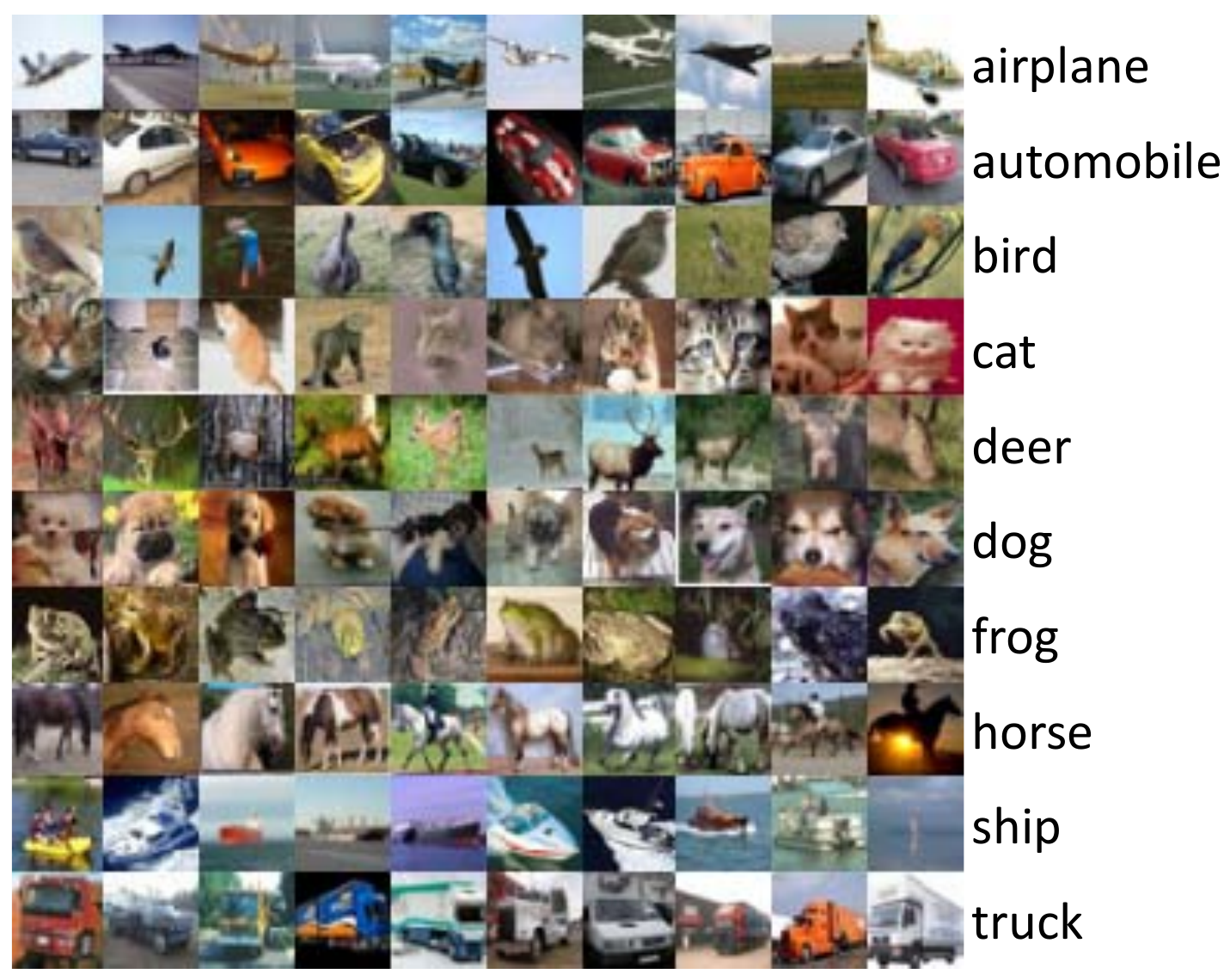}}
\end{center}
   \caption{The sample images from CIFAR-10.}
\label{fig:1}
\end{figure}

\begin{figure}
\begin{center}
\scalebox{0.5}{\includegraphics{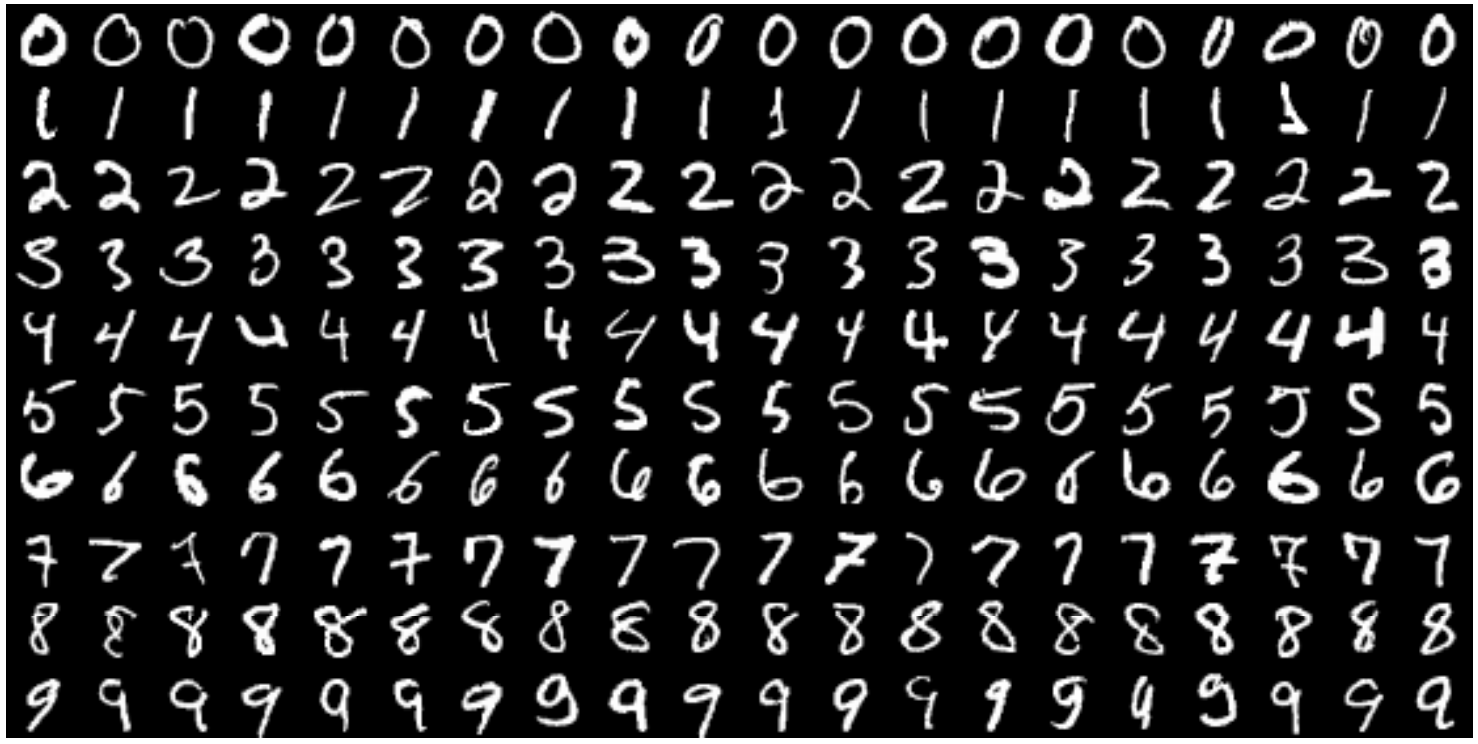}}
\end{center}
   \caption{The sample images on the MNIST database.}
\label{fig:2}
\end{figure}

As a subset of the famous 80M tiny image collection \cite{torralba200880}, CIFAR-10 contains 60,000 images from 10 classes with 6,000 examples per class. Example images from this dataset are presented in Fig. \ref{fig:1}. Each image in this dataset is represented as a 512-dimensional GIST feature vector \cite{oliva2001modeling}. MNIST contains 70,000 784-dimensional handwritten digit images from `0' to `9', each image being 28$\times$28 pixels; some example images are shown in Fig. \ref{fig:2}. Both MNIST and CIFAR-10 are split into a test set with 1,000 examples and a training set containing all remaining examples. FRGC is described below.
\subsection{Experiments on CIFAR-10}
\begin{figure}
\begin{center}
\scalebox{0.45}{\includegraphics{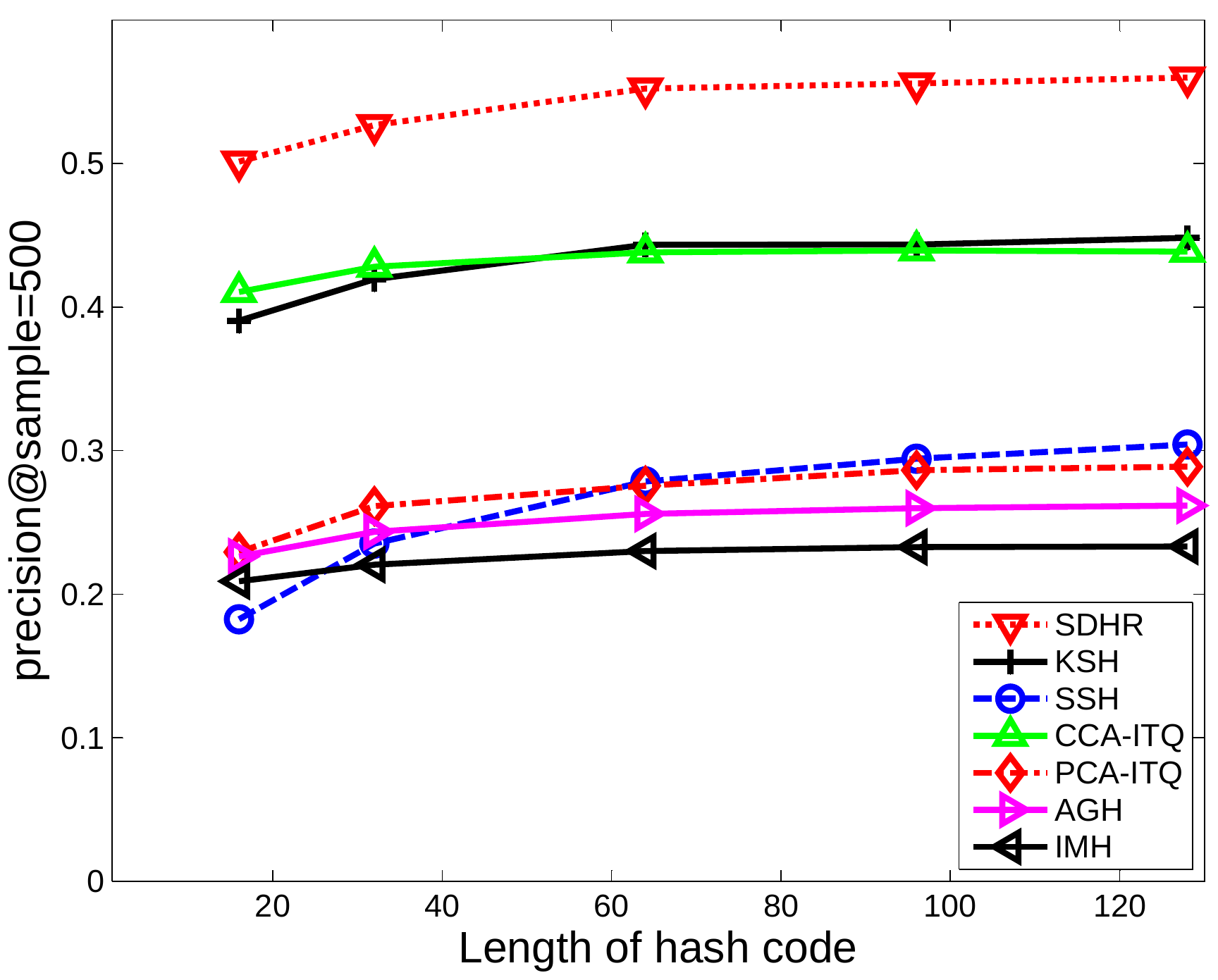}}
\end{center}
   \caption{Precision@sample=500 versus the number of hashing bits (16, 32, 64, 96, 128) on CIFAR-10.}
\label{fig:3}
\end{figure}

\begin{figure}
\begin{center}
\scalebox{0.45}{\includegraphics{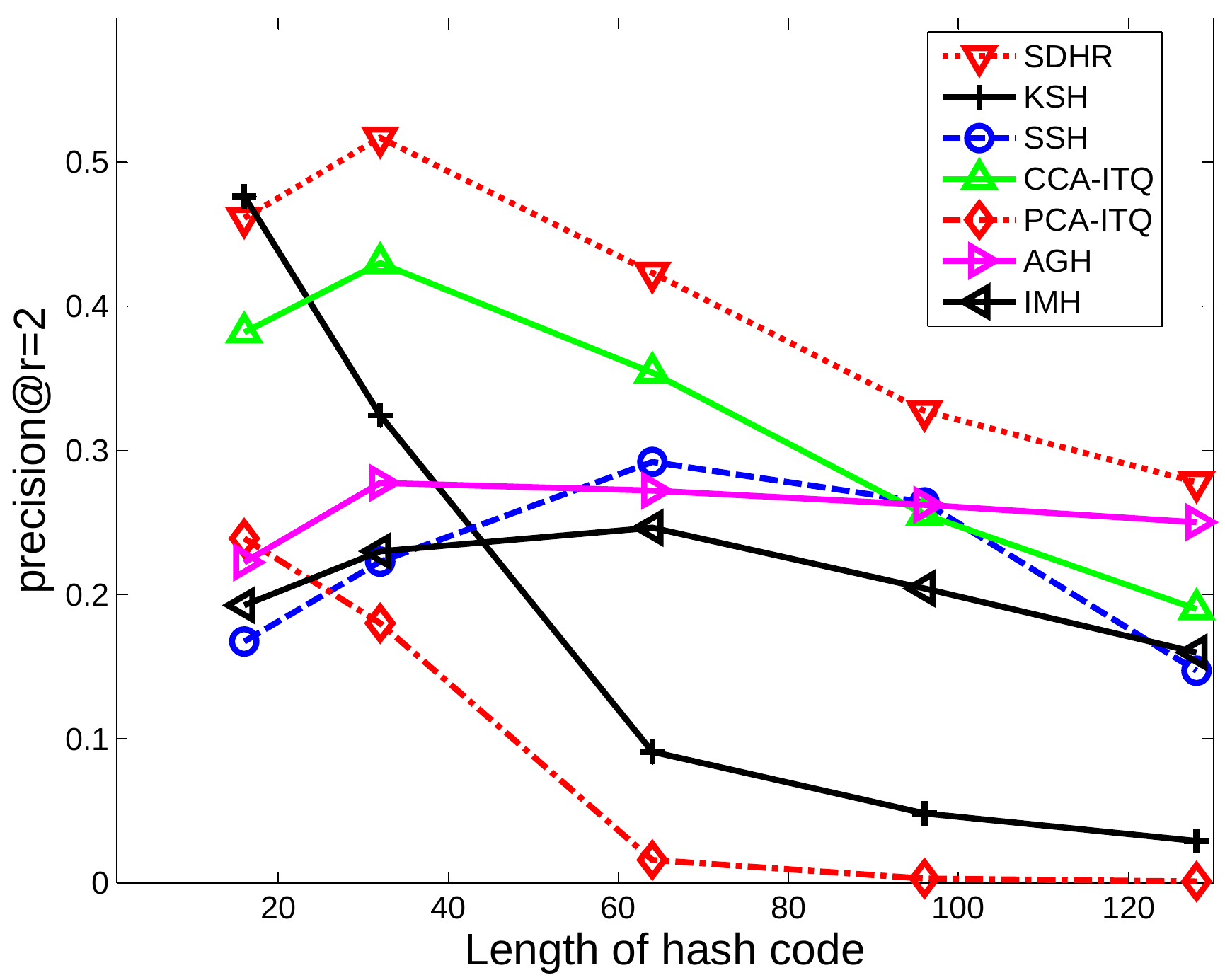}}
\end{center}
   \caption{Precision of Hamming radius 2 versus the number of hashing bits (16, 32, 64, 96, 128) on CIFAR-10.}
\label{fig:4}
\end{figure}

\begin{figure}
\begin{center}
\scalebox{0.45}{\includegraphics{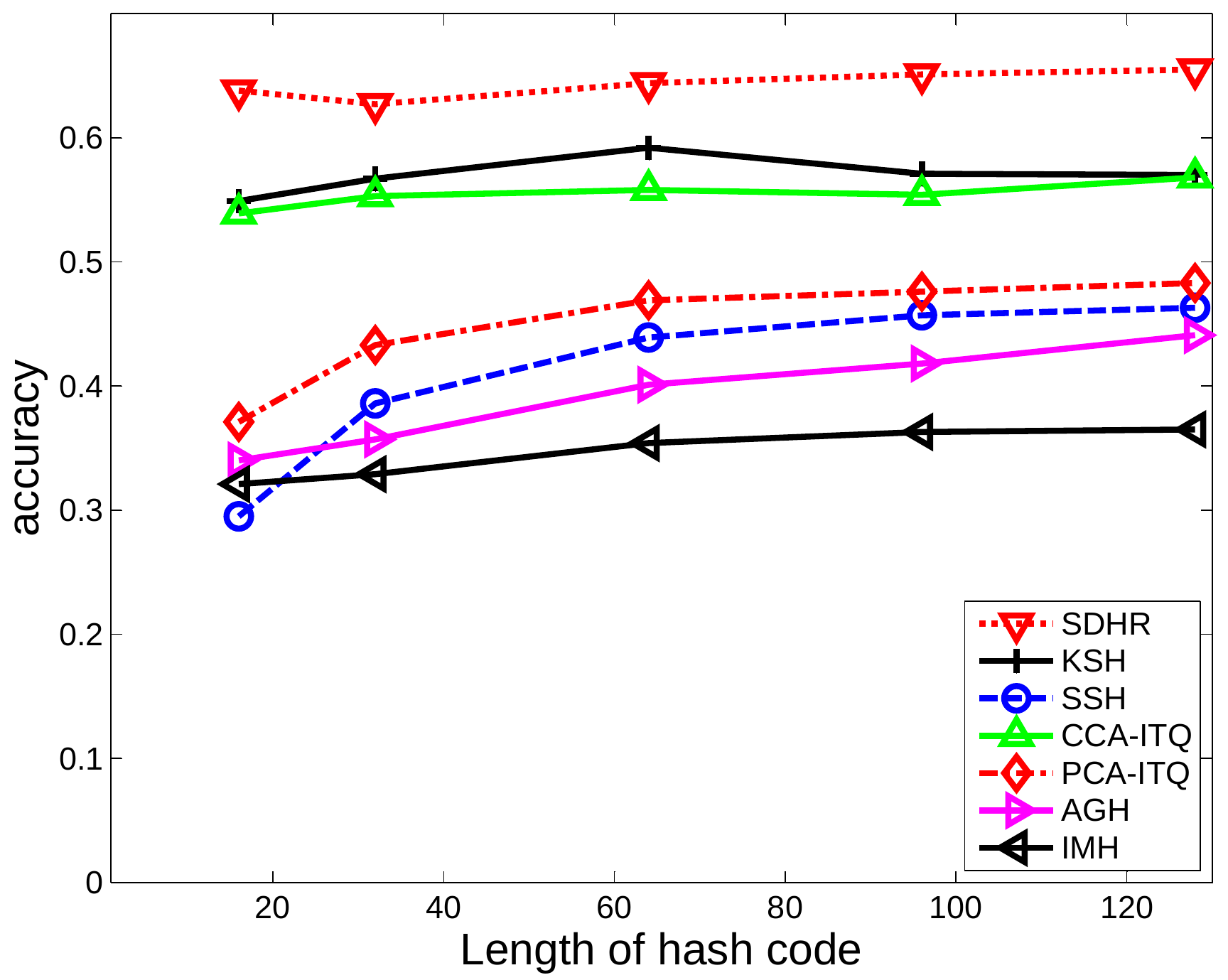}}
\end{center}
   \caption{Accuracy versus the number of hashing bits (16, 32, 64, 96, 128) on CIFAR-10.}
\label{fig:5}
\end{figure}

The experimental results on CIFAR-10 are shown in Table \ref{tab:1}. SDHR performs better than SDH in terms of precision, recall, F-measure, and accuracy. For example, the accuracy of SDHR is 1.03-times higher than that of SDH. Moreover, from the last two columns in Table \ref{tab:1}, we can see that the training time cost of SDHR is lower than that of SDH, which allows the method to be applied to the whole training data. In contrast, KSH and FastHash take about 20 minutes to train. More specifically, the training of SDHR is about 28-times and 31-times faster than FastHash and KSH, respectively, in this case. SSH, CCA-ITQ, PCA-ITQ, AGH, and IMH are also very efficient; however, their performance is generally worse than SDHR. The precision at 500 examples (precision@sample = 500), precision of Hamming radius 2, and accuracy versus the number of hashing bits are presented in Figs. \ref{fig:3}-\ref{fig:5}, respectively. Due to space limitations, we only show some algorithms in the corresponding figure. With respect to precision of Hamming radius 2, SDHR outperforms the other methods when the number of hashing bits is larger than 32, and KSH performs the best when the number of hashing bits is 16. SDHR outperforms all other methods in terms of precision at 500 examples (precision@sample = 500) and accuracy, highlighting the effectiveness of SDHR.
\begin{table*}
\begin{center}
\scalebox{1}[1]{
\begin{tabular}{|l|c|c|c|c|c|c|c|}
\hline
Method & precision@$r$=2 & recall@$r$=2 & F-measure@$r$=2 & MAP & accuracy & training time & test time\\
\hline\hline
SDHR & 0.4612  & 0.2835  &  \textbf{0.3511} &  0.4091 & \textbf{0.638}  &  42.4 & 2.7e-6  \\
SDH  & 0.4588  & 0.2805  &  0.3481 &  0.4123 & 0.622  & 43.9  &  2.2e-6 \\
BRE  & 0.1745  & 0.2118  &  0.1913 &  0.1431 & 0.305  & 66.0  &  6.3e-6 \\
KSH  & 0.4763  & 0.1167  &  0.1875 &  0.4030 & 0.549  & 1.3e3 &  7.8e-5  \\
SSH  & 0.1674  & \textbf{0.3908}  & 0.2344  & 0.1687  & 0.295  & 17.2  & 3.0e-6  \\
CCA-ITQ  & 0.3819  &  0.1382 & 0.2030  & 0.3137  & 0.539  & 4.6  & \textbf{1.5e-7}  \\
FastHash  & \textbf{0.5560}  & 0.2310  & 0.3264  & \textbf{0.5253}  & 0.588  & 1.2e3  & 4.3e-4  \\
PCA-ITQ  &  0.2389 &  0.0221 & 0.0404  & 0.1616  &  0.371 &  \textbf{3.3} &  \textbf{1.5e-7} \\
AGH  & 0.2223  &  0.0553 & 0.0886  & 0.1562  &  0.34 &  6.9 &  8.5e-5 \\
IMH  & 0.1926  &  0.1454 & 0.1657  & 0.1696  &  0.321&  42.5 & 6.3e-5  \\
\hline
\end{tabular}}
\end{center}
\caption{Precision, recall, F-measure of Hamming distance within radius 2, MAP, accuracy, and time on CIFAR-10. Results are reported when the number of hashing bits is 16. For SSH, 5,000 labeled examples are used for similarity matrix construction. The training and test times are in seconds.}\label{tab:1}
\end{table*}

\subsection{Experiments on MNIST}
\begin{figure}
\begin{center}
\scalebox{0.45}{\includegraphics{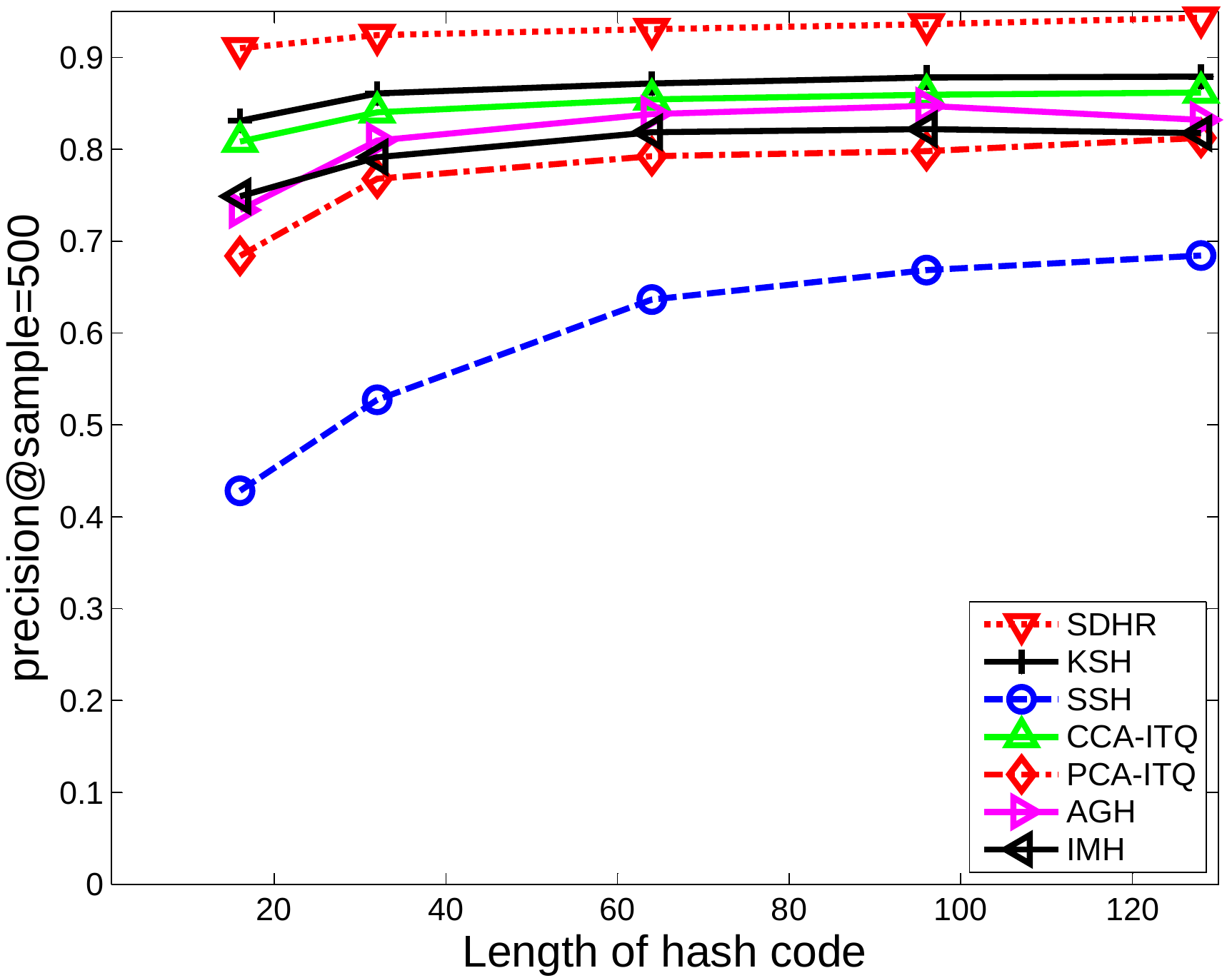}}
\end{center}
   \caption{Precision@sample=500 versus the number of hashing bits (16, 32, 64, 96, 128) on MNIST.}
\label{fig:6}
\end{figure}

\begin{figure}
\begin{center}
\scalebox{0.45}{\includegraphics{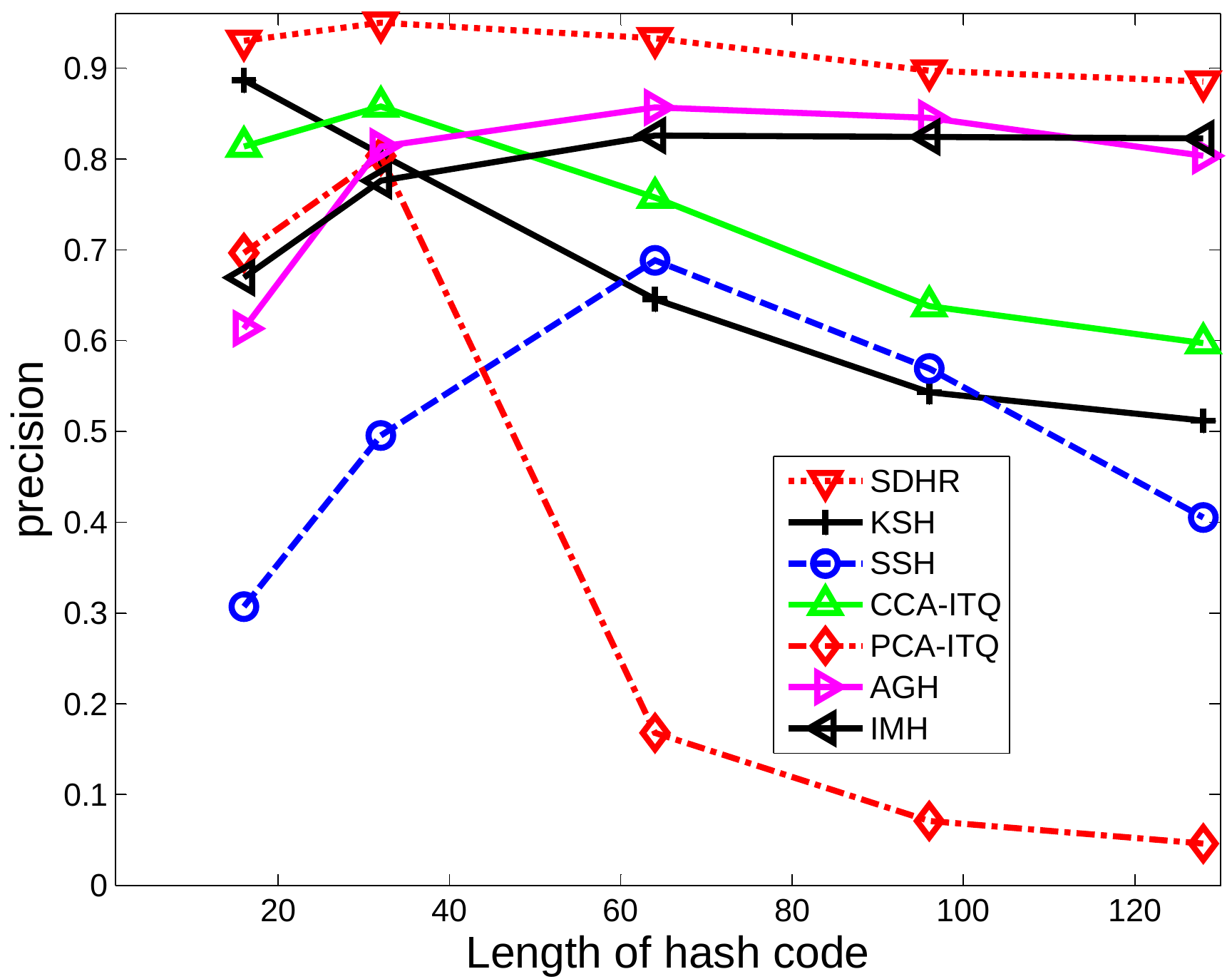}}
\end{center}
   \caption{Precision of Hamming radius 2 versus the number of hashing bits (16, 32, 64, 96, 128) on MNIST.}
\label{fig:7}
\end{figure}

\begin{figure}
\begin{center}
\scalebox{0.45}{\includegraphics{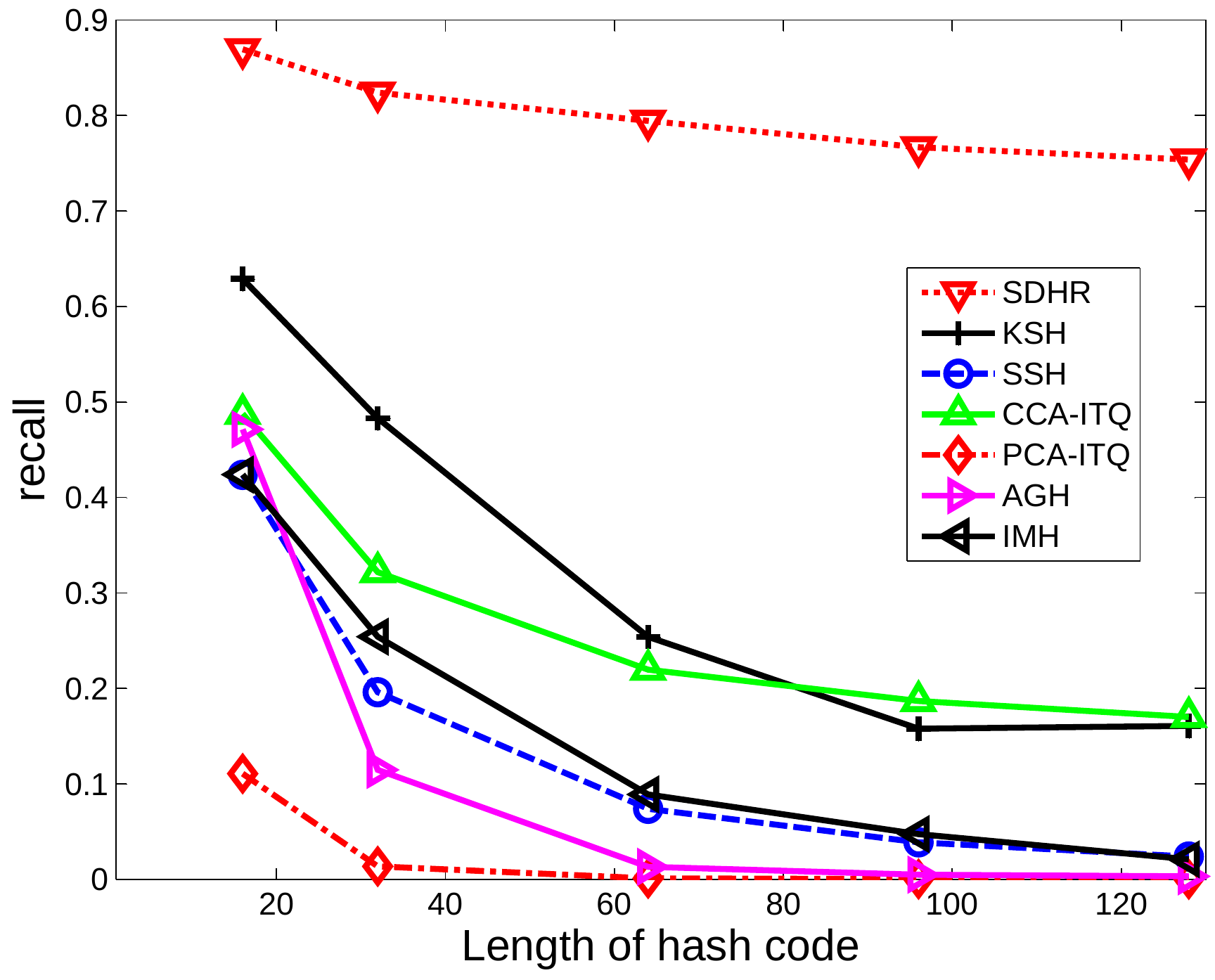}}
\end{center}
   \caption{Recall of Hamming radius 2 versus the number of hashing bits (16, 32, 64, 96, 128) on MNIST.}
\label{fig:8}
\end{figure}

\begin{figure}
\begin{center}
\scalebox{0.45}{\includegraphics{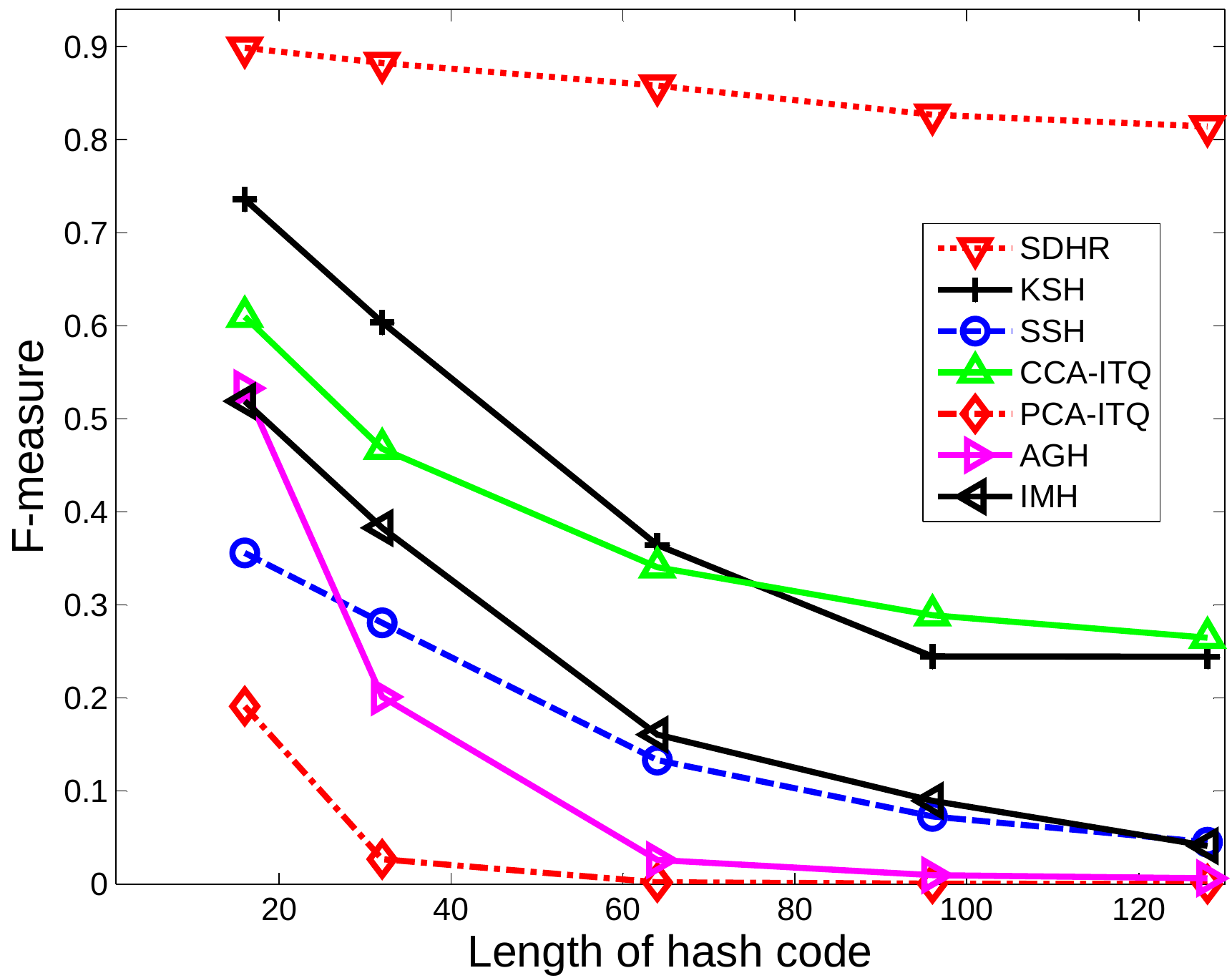}}
\end{center}
   \caption{F-measure of Hamming radius 2 versus the number of hashing bits (16, 32, 64, 96, 128) on MNIST.}
\label{fig:9}
\end{figure}

\begin{figure}
\begin{center}
\scalebox{0.45}{\includegraphics{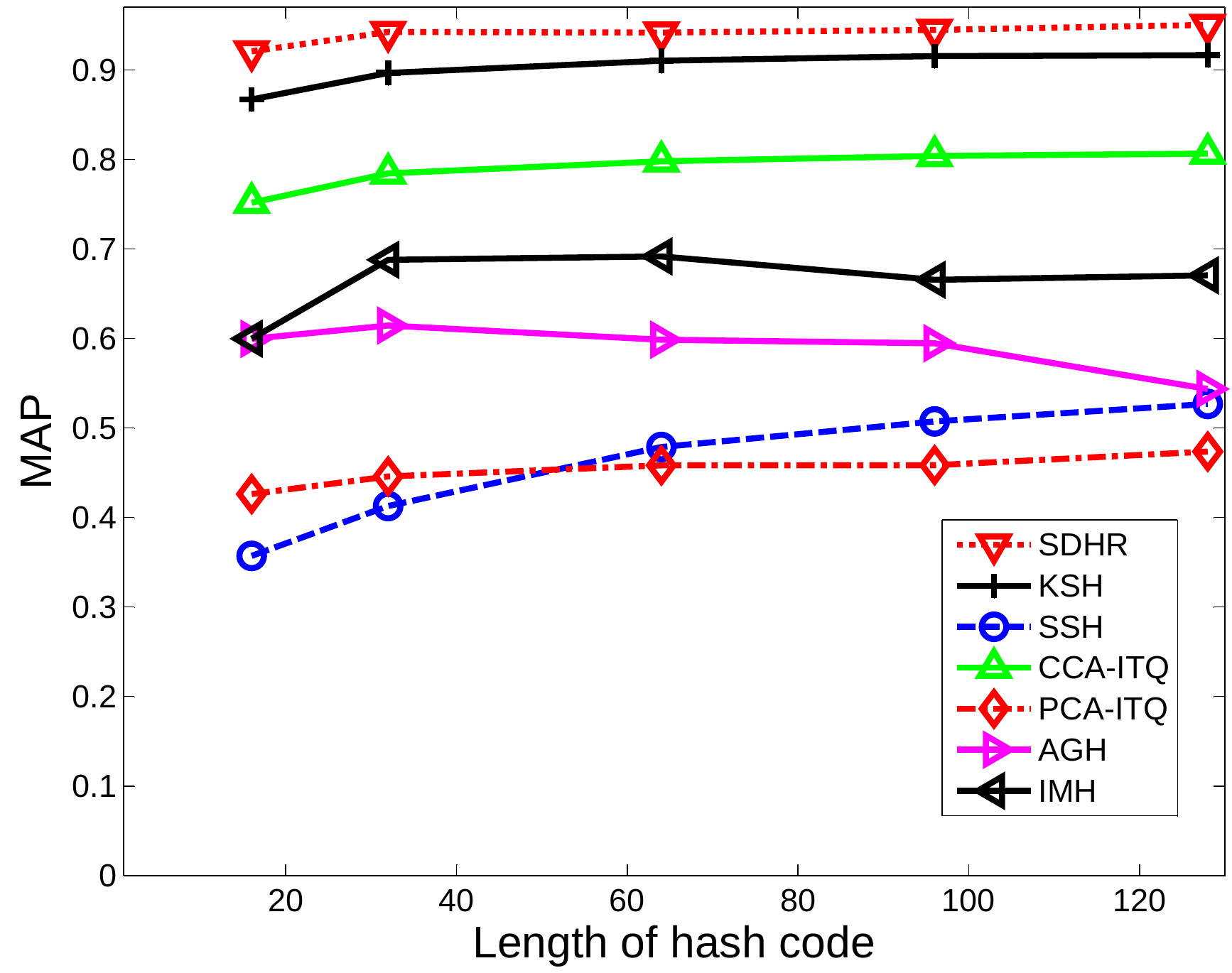}}
\end{center}
   \caption{MAP versus the number of hashing bits (16, 32, 64, 96, 128) on MNIST.}
\label{fig:10}
\end{figure}

\begin{figure}
\begin{center}
\scalebox{0.45}{\includegraphics{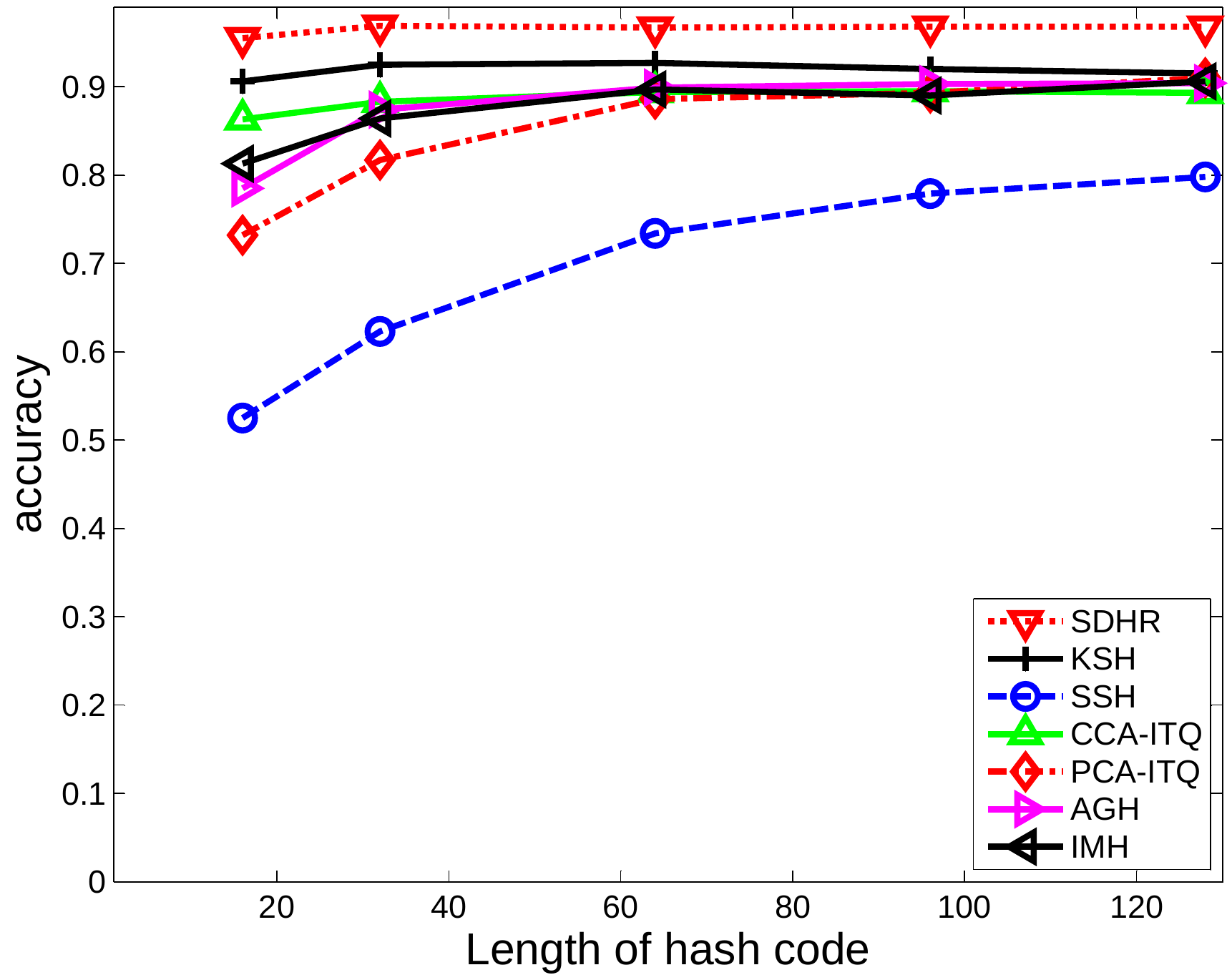}}
\end{center}
   \caption{Accuracy versus the number of hashing bits (16, 32, 64, 96, 128) on MNIST.}
\label{fig:11}
\end{figure}

The experimental results on MNIST are presented in Table \ref{tab:2}. SDHR performs best in terms of precision, recall, and F-measure, while FastHash performs best in terms of MAP and accuracy. However, SDHR is much faster than FastHash: SDHR and FastHash take 186.2 and 4661.1 seconds, respectively. Thus, SDHR is about 25-times faster than FastHash in this case. Furthermore, the precision@sample=500, precision of Hamming radius 2, recall of Hamming radius 2, F-measure of Hamming radius 2, MAP, and accuracy curves are shown in Figs. \ref{fig:6}-\ref{fig:11}, respectively. Only some methods are shown due to space limitations. SDHR outperforms the other methods and, furthermore, SDHR particularly outperforms the other methods in terms of recall of Hamming radius 2 and F-measure of Hamming radius 2.
\begin{table*}
\begin{center}
\scalebox{1}[1]{
\begin{tabular}{|l|c|c|c|c|c|c|c|}
\hline
Method & precision@$r$=2 & recall@$r$=2 & F-measure@$r$=2 & MAP & accuracy & training time & test time\\
\hline\hline
SDHR &  \textbf{0.9330} &  \textbf{0.7943} & \textbf{0.8581}  & 0.9417  & 0.967  & 186.2  &  4.9e-6 \\
SDH  &  0.9269 & 0.7711  & 0.8419  & 0.9397  &  0.963 & 128  & 5.1e-6  \\
BRE  & 0.3850  & 0.0011  & 0.0021  & 0.4211  &  0.839 & 24060.6  & 9.3e-5  \\
KSH  & 0.6454  & 0.2539  & 0.3644  & 0.9103  &  0.927 & 1324.7  &  7.6e-5 \\
SSH  & 0.6883  & 0.0738  & 0.1332  & 0.4787  &  0.734 & 260.2  &  5.7e-6 \\
CCA-ITQ  & 0.7575  & 0.2196  & 0.3405  & 0.7978  & 0.894  &  16.6 &  \textbf{4.1e-7} \\
FastHash  &0.8680   & 0.6735  &  0.7585 &  \textbf{0.9813} &  \textbf{0.972} & 4661.1  & 0.0012  \\
PCA-ITQ  &  0.1680 & 9.3e-4  & 0.0018  & 0.4581  & 0.886  & 10.1  & 4.5e-7  \\
AGH  &  0.8568 & 0.0131  & 0.0258  & 0.5984  & 0.899  & \textbf{6.9}  & 6.4e-5  \\
IMH  &  0.8258 & 0.0889  &  0.1606 & 0.6916  &  0.897 & 32.2  &  6.4e-5 \\
\hline
\end{tabular}}
\end{center}
\caption{Precision, recall, F-measure of Hamming distance within radius 2, MAP, accuracy, training time, and test time on MNIST. Results are reported when the number of hashing bits is 64. }\label{tab:2}
\end{table*}

\subsection{Experiments on FRGC}
\begin{figure}
\begin{center}
\scalebox{0.5}{\includegraphics{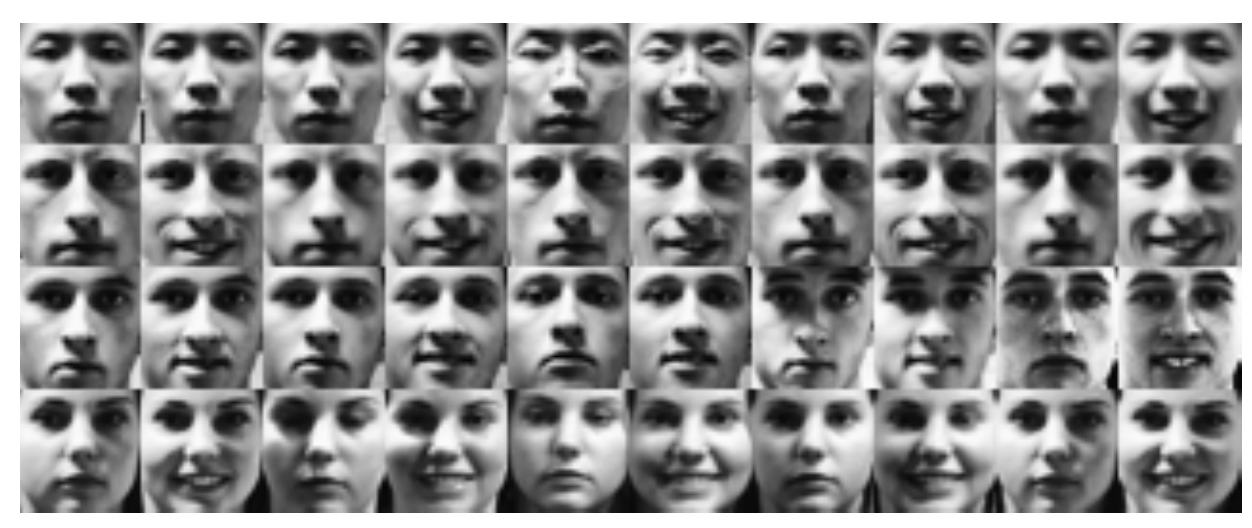}}
\end{center}
   \caption{Cropped and resized examples of four randomly selected individuals in the FRGC face dataset. Each row represents one individual.}
\label{fig:12}
\end{figure}

The FRGC version two face dataset \cite{phillips2005overview} is a large-scale and challenging benchmark face dataset. FRGC experiment 4 contains 8014 face images from 466 persons in the query set. These uncontrolled images demonstrate variations in blur, expression, illumination, and time. In our experiment, we only select persons with over 10 images in the dataset, resulting in 3160 images from 316 persons. Each image is cropped and resized to 32$\times$32 pixels (256 gray levels per pixel) by fixing the positions of the eyes. For each individual, three images are randomly selected for testing and the remaining seven used for training. Fig. \ref{fig:12} shows example images from FRGC.

\begin{figure}
\begin{center}
\scalebox{0.45}{\includegraphics{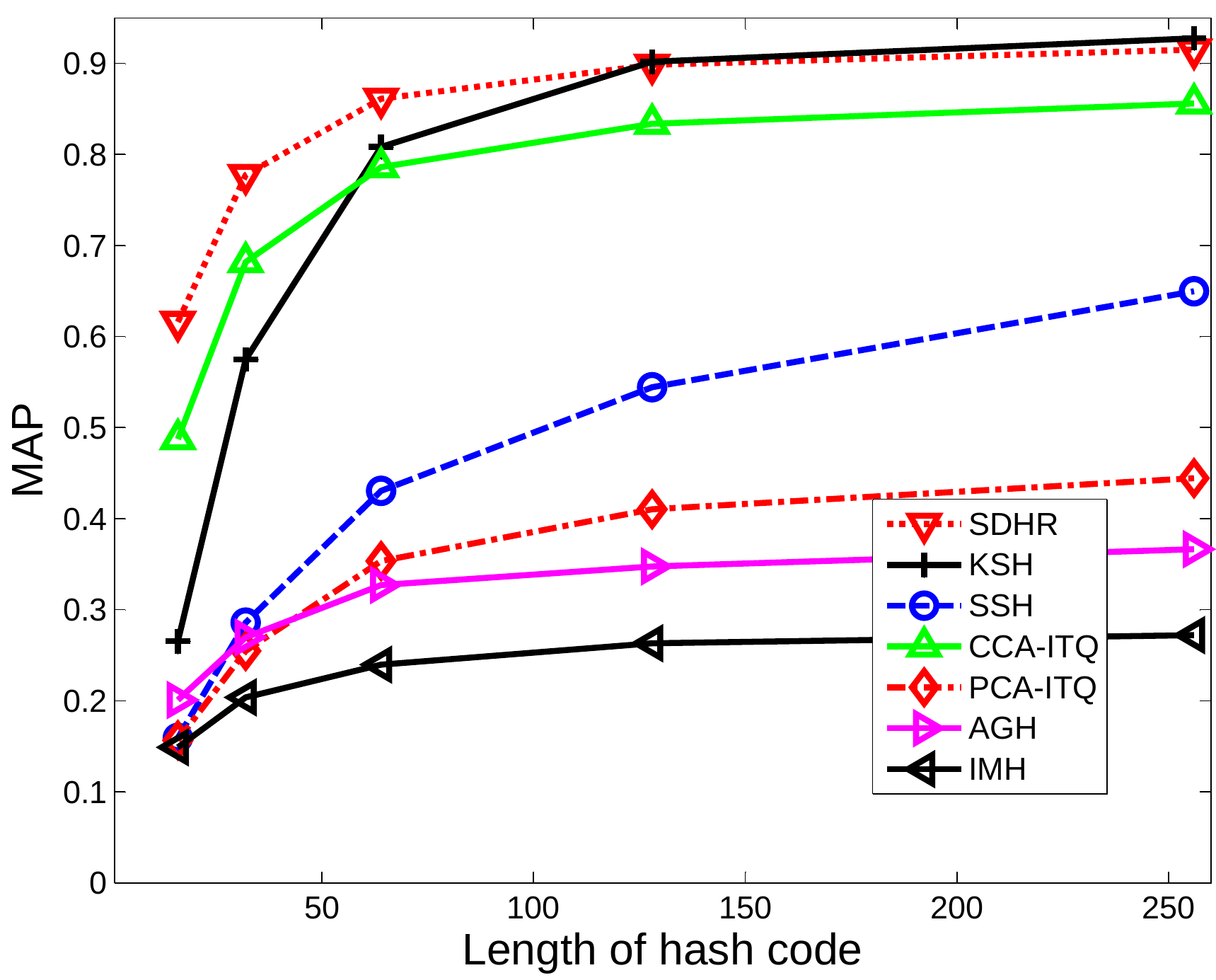}}
\end{center}
   \caption{MAP versus the number of hashing bits (16, 32, 64, 128, 256) on FRGC.}
\label{fig:13}
\end{figure}

\begin{figure}
\begin{center}
\scalebox{0.45}{\includegraphics{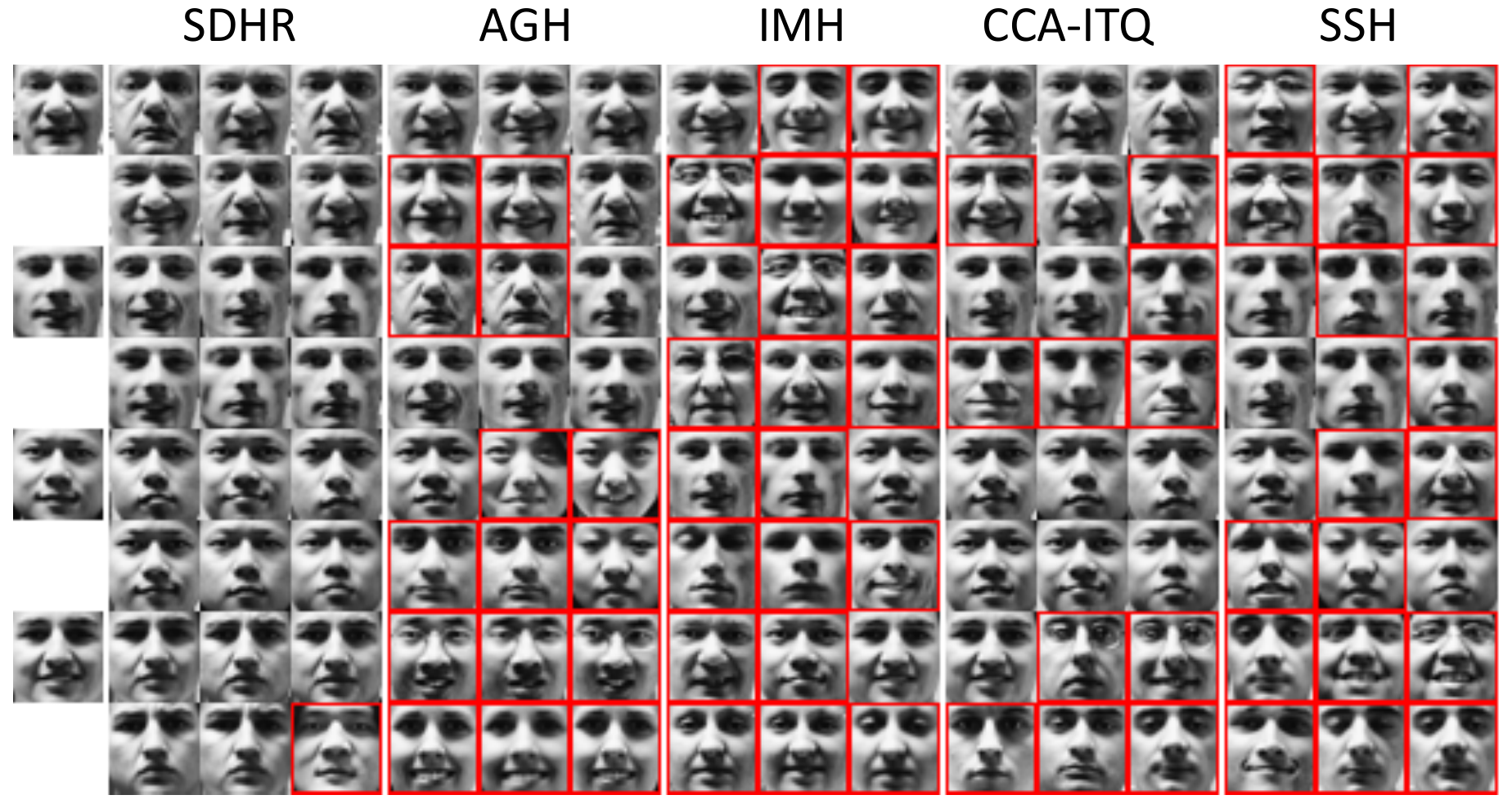}}
\end{center}
   \caption{Top 6 retrieved images of 4 queries returned by various hashing algorithms on FRGC. The query image is in the first column. From left to right: retrieved images by SDHR, AGH, IMH, CCA-ITQ, and SSH when 16-bit binary codes are used for searching. False positive returns are marked with red borders.}
\label{fig:14}
\end{figure}

The experimental results on FRGC are presented in Table \ref{tab:3}. SDHR outperforms the other methods in terms of precision, recall, F-measure, and MAP, while CCA-ITQ outperforms the other methods in terms of accuracy. Fig. \ref{fig:13} shows MAP versus the number of hashing bits on this dataset. Due to space limitations, Fig. \ref{fig:13} only shows representative methods. This figure illustrates two main points: first, SDHR outperforms other methods when the number of hashing bits is less than or equals 64, while KSH performs best when the number of hashing bits is larger than or equal to 128; second, the MAP of all methods increases as the number of hashing bits increases. This might be because, as the number of hashing bits increases, more information can be encoded in the hash code. Therefore, the hash code represents the face image in a more informative and discriminative way. Furthermore, Fig. \ref{fig:14} shows some example query images and the retrieved neighbors on FRGC when 16 bits are used to learn the hash codes. We can see that SDHR shows better searching performance because higher semantic relevance is obtained in the top retrieved examples.

\begin{table*}
\begin{center}
\scalebox{1}[1]{
\begin{tabular}{|l|c|c|c|c|c|c|c|}
\hline
Method & precision@$r$=2 & recall@$r$=2 & F-measure@$r$=2 & MAP & accuracy & training time & test time\\
\hline\hline
SDHR & \textbf{0.4893}  & \textbf{0.5893}  & \textbf{0.5346}  & \textbf{0.6161}  & 0.563  & 5.9  &  1.4e-6 \\
SDH  &  0.4851 & 0.5833  & 0.5297  &  0.6138 &  0.565 & 1.6  & 1.7e-6  \\
BRE  &  0.0352 & 0.3798  &  0.0645 &  0.0851 &  0.145 &  117.7 & 1.3e-5  \\
KSH  &  0.2697 & 0.2400  &  0.2540 &  0.2654 &  0.377 &  333.1 & 1.1e-4  \\
SSH  &  0.1286 & 0.2402  &  0.1675 &  0.1593 &  0.263 &  5.8 & 5.7e-6  \\
CCA-ITQ  & 0.4426  &  0.4776 & 0.4594  &  0.4874 &  \textbf{0.59} & 1.3  &  1.1e-7 \\
FastHash  & 0.1646  & 0.0895  & 0.1160  &  0.1199 & 0.283  & 62.0  & 6.2e-4  \\
PCA-ITQ  & 0.1359  &  0.2307 &  0.1710 &  0.1566 & 0.275  &  \textbf{0.2} & \textbf{9.7e-8}  \\
AGH  & 0.0688  &  0.4107 &  0.1179 & 0.2009  & 0.237  & 2.3  &  1.2e-4 \\
IMH  & 0.0492  & 0.4460  & 0.0886  & 0.1489  & 0.198  & 35.0  & 8.9e-5  \\
\hline
\end{tabular}}
\end{center}
\caption{Precision, recall, F-measure of Hamming distance within radius 2, MAP, accuracy and time on FRGC. Results are reported when the number of hashing bits is 16. }\label{tab:3}
\end{table*}

\section{Conclusions}
In this paper, we propose a novel learning-based hashing method called ``Supervised Discrete Hashing with Relaxation" (SDHR) based on ``Supervised Discrete Hashing" (SDH). SDHR uses learned code words rather than the traditional fixed code words used in SDH to encode class label information. As expected, the SDHR's performance is better than that of SDH. Real-world image classification and face recognition experiments highlight the advantages of the proposed method.

Although hashing methods can reduce computational costs, the costs remain very large for large-scale image retrieval. Selecting or learning representative examples to represent each class must be further studied.

We used the pixel feature on the MNIST and FRGC datasets. It is possible that advanced visual features rather than the original pixel intensity values will further improve performance. Designing the most appropriate feature for specific hashing algorithms is also a thought-provoking direction for future studies.

\ifCLASSOPTIONcaptionsoff
  \newpage
\fi



%
\bibliographystyle{ieeetr}
\bibliography{a}

%
\begin{IEEEbiography}[{\includegraphics[width=1in,height=1.25in,clip,keepaspectratio]{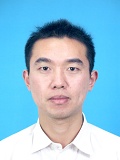}}]{Jie Gui} (SM'16) is an associate professor in Hefei Institute of Intelligent Machines, Chinese Academy of Sciences. He received the BS degree in Computer Science from the Hohai University, Nanjing, China, in 2004, the MS degree in Computer Applied Technology from Hefei Institutes of Physical Science, Chinese Academy of Sciences, Hefei, China, in 2007, and the PhD degree in Pattern Recognition and Intelligent Systems from the University of Science and Technology of China, Hefei, China, in 2010. He has been a postdoctoral fellow in the National Laboratory of Pattern Recognition, Institute of Automation Chinese Academy of Sciences. His research interests include machine learning, pattern recognition, data mining and image processing.
\end{IEEEbiography}

\begin{IEEEbiography}[{\includegraphics[width=1in,height=1.25in,clip,keepaspectratio]{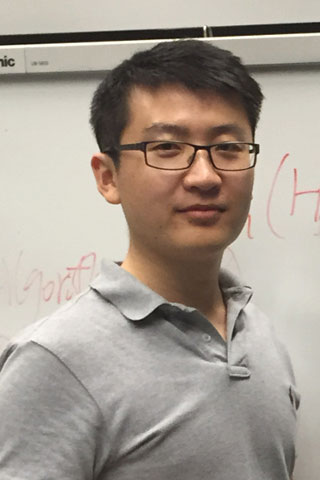}}]{Tongliang Liu}
received the BE degree in electronic engineering and information science from the University of Science and Technology of China, Hefei, China, in 2012, and the PhD degree from the University of Technology Sydney, Sydney, Australia, in 2016. He is currently a Lecturer with the Faculty of Engineering and Information Technology in the University of Technology Sydney. His research interests include statistical learning theory, computer vision, and optimization. He has authored and co-authored 10+ research papers including IEEE T-PAMI, T-NNLS, T-IP, NECO, ICML, KDD, IJCAI, and AAAI. He won the Best Paper Award in IEEE International Conference on Information Science and Technology 2014.
\end{IEEEbiography}

\begin{IEEEbiography}[{\includegraphics[width=1in,height=1.25in,clip,keepaspectratio]{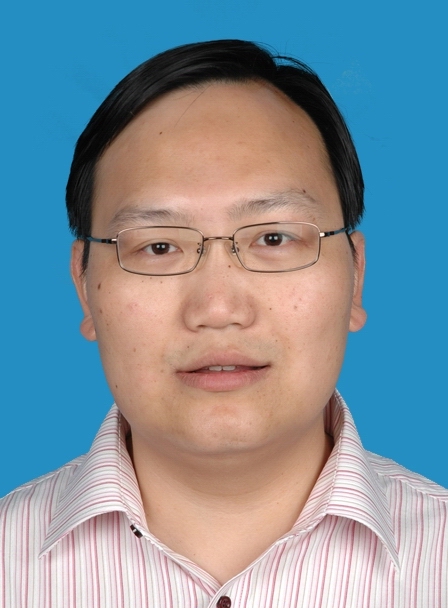}}]{Zhenan Sun}
(M'07) is currently a Professor in National Laboratory of Pattern Recognition (NLPR) at Institute of Automation, Chinese Academy of Sciences (CASIA). He received the BS degree in Industrial Automation from Dalian University of Technology, the MS degree in System Engineering from Huazhong University of Science and Technology, and the PhD degree in Pattern Recognition and Intelligent Systems from CASIA in 1999, 2002 and 2006, respectively. Since March 2006, Dr. Sun has joined the NLPR of CASIA as a faculty member. Dr. Sun is a member of the IEEE, the IEEE Computer Society and the IEEE Signal Processing Society. His research areas include biometrics, pattern recognition and computer vision, and he has authored/co-authored over 100 technical papers. He is an associated editor of IEEE Transactions on Information Forensics and Security and IEEE Biometrics Compendium.
\end{IEEEbiography}

\begin{IEEEbiography}[{\includegraphics[width=1in,height=1.25in,clip,keepaspectratio]{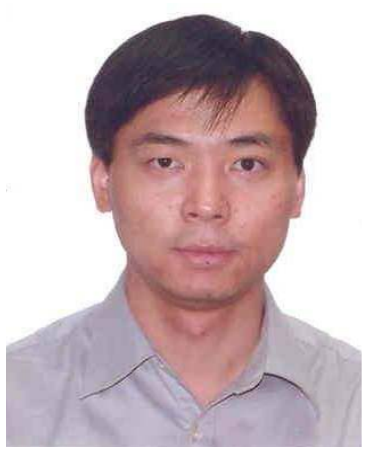}}]{Dacheng Tao}
 (F'15) is Professor of School of Information Technologies, Faculty of Engineering and Information Technologies, University of Sydney, NSW, Australia. He mainly applies statistics and mathematics to data analytics problems and his research interests spread across computer vision, data science, image processing, machine learning, and video surveillance. His research results have expounded in one monograph and 200+ publications at prestigious journals and prominent conferences, such as IEEE T-PAMI, T-NNLS, T-IP, JMLR, IJCV, NIPS, ICML, CVPR, ICCV, ECCV, AISTATS, ICDM; and ACM SIGKDD, with several best paper awards, such as the best theory/algorithm paper runner up award in IEEE ICDM'07, the best student paper award in IEEE ICDM'13, and the 2014 ICDM 10-year highest-impact paper award. He received the 2015 Australian Scopus-Eureka Prize, the 2015 ACS Gold Disruptor Award and the 2015 UTS Vice-Chancellor¡¯s Medal for Exceptional Research. He is a Fellow of the IEEE, OSA, IAPR and SPIE.
\end{IEEEbiography}

\begin{IEEEbiography}[{\includegraphics[width=1in,height=1.25in,clip,keepaspectratio]{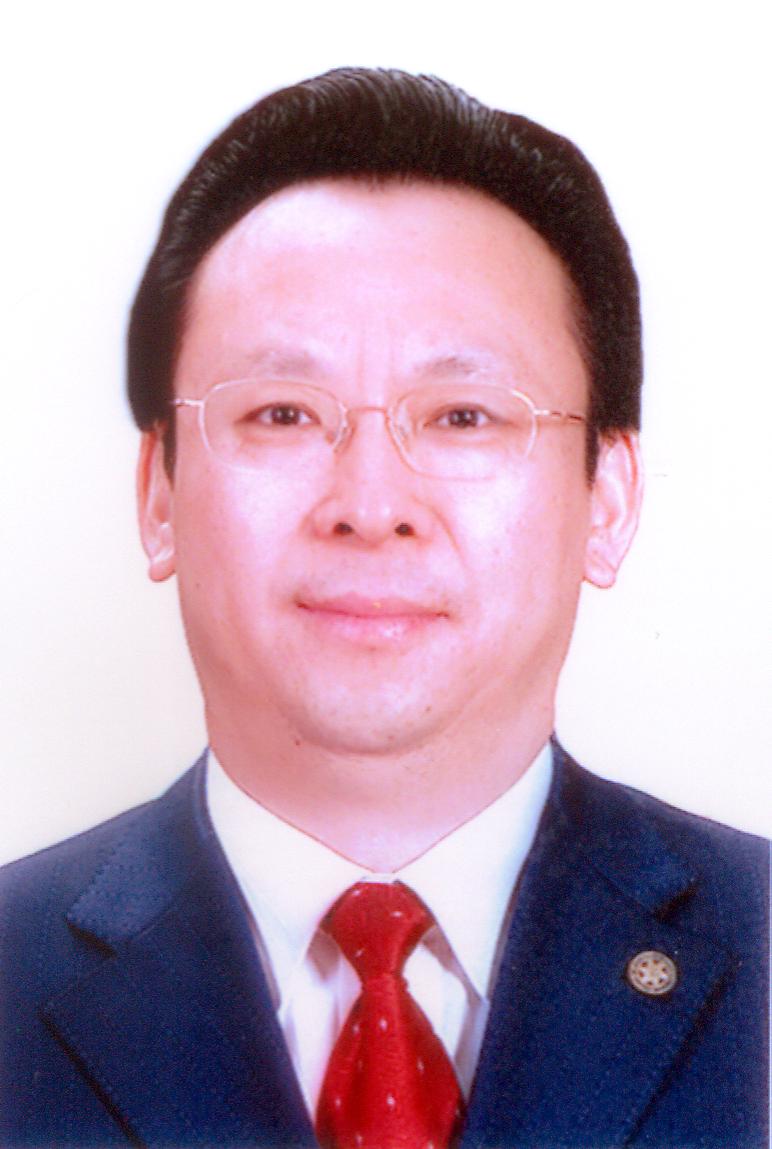}}]{Tieniu Tan}
(F'03) received his BS degree in electronic engineering from Xi'an Jiaotong University, China, in 1984, and his MS and PhD degrees in electronic engineering from Imperial College London, U.K., in 1986 and 1989, respectively.

In October 1989, he joined the Department of Computer Science, the University of Reading, U.K., where he worked as a Research Fellow, Senior Research Fellow and Lecturer. In January 1998, he returned to China to join the National Laboratory of Pattern Recognition (NLPR), Institute of Automation of the Chinese Academy of Sciences (CAS) as a full professor. He was the Director General of the CAS Institute of Automation from 2000-2007, and the Director of the NLPR from 1998-2013. He is currently Director of the Center for Research on Intelligent Perception and Computing at the Institute of Automation and also serves as Deputy President of the CAS. He has published more than 500 research papers in refereed international journals and conferences in the areas of image processing, computer vision and pattern recognition, and has authored or edited 11 books. He holds more than 70 patents. His current research interests include biometrics, image and video understanding, and information forensics and security.

Dr Tan is a Member (Academician) of the Chinese Academy of Sciences, Fellow of The World Academy of Sciences for the advancement of sciences in developing countries (TWAS), an International Fellow of the UK Royal Academy of Engineering, and a Fellow of the IEEE and the IAPR (the International Association of Pattern Recognition). He is Editor-in-Chief of the International Journal of Automation and Computing. He has given invited talks and keynotes at many universities and international conferences, and has received numerous national and international awards and recognitions.
\end{IEEEbiography}






\end{document}